\documentclass[preprint,12pt]{elsarticle}

\usepackage{amssymb}

\usepackage{amsmath,amssymb,amsfonts}
\usepackage{algorithmic}
\usepackage{graphicx}
\usepackage{textcomp}

\usepackage{subfigure}
\usepackage{multirow}
\usepackage{color}
\usepackage[table]{xcolor}
\usepackage{hyperref}
\usepackage{booktabs}
\usepackage{dsfont}
\usepackage{threeparttable}

\begin{document}

\begin{frontmatter}



\title{Discriminative-Generative Synergy for Occlusion Robust 3D Human Mesh Recovery}


\author[label1]{Yang Liu}

\author[label1]{Zhiyong Zhang}

\address[label1]{School of Electronics and Communication Engineering, Sun Yat-sen University, Shenzhen, Guangdong, China}

\begin{abstract}
	3D human mesh recovery from monocular RGB images is a fundamental yet challenging task, aiming to estimate anatomically plausible 3D model parameters that provide rich geometric and semantic representations for downstream applications. While mainstream regression-based paradigms excel in inference speed, they severely struggle with partial and severe occlusions, often leading to physically implausible or geometrically erroneous estimations in unconstrained real-world scenarios. Conversely, recent denoising diffusion probabilistic models offer powerful generative priors capable of plausibly hallucinating occluded regions. However, their inherently generative nature may override weak visual conditioning when confronted with rare poses, failing to faithfully reconstruct the authentic poses. To overcome the limitations of existing single-paradigm approaches, we propose a novel brain-inspired synergistic framework that intrinsically integrates the robust discriminative representation capacity of vision transformers with the powerful generative priors of conditional diffusion models. Analogous to the left and right hemispheres of the human brain, the ViT-centric discriminative pathway acts as logical deduction to extract deterministic visual cues from visible regions, while the generative pathway leverages diffusion models to synthesize structurally coherent body structures. Furthermore, to effectively bridge these two pathways, we introduce a diverse-consistent feature learning module to align the discriminative representations with generative priors, and a cross-attention multi-level fusion mechanism to facilitate bidirectional feature interaction across multiple semantic levels. Extensive experiments on standard benchmark datasets demonstrate that our approach achieves superior performance across key evaluation metrics, validating its effectiveness and enhanced robustness in complex, real-world scenarios.
\end{abstract}



\begin{keyword}


	3D human mesh recovery\sep
	Vision transformer\sep
	Diffusion model\sep
	Occlusion\sep
	Diverse-consistent feature learning

\end{keyword}

\end{frontmatter}



\section{Introduction} \label{sec1}

	Driven by rapid advancements in virtual reality, augmented reality, digital twins, and human-computer interaction, the accurate and robust 3D digitization of humans in real-world scenarios has emerged as a pivotal technology \cite{zhu2024bed, zhang2024virtual, sun2025dynamic}. 
	In this context, 3D human mesh recovery from monocular RGB images has become a fundamental yet challenging task within the computer vision community. 
	This task aims to estimate anatomically plausible 3D model parameters that are accurately aligned with the imaged subjects. 
	By capturing human poses, body shapes, and even subtle facial expressions and hand gestures, it provides rich geometric and semantic representations for downstream applications, thereby pushing the boundaries of existing technologies.

\begin{figure}[htbp]
		\centering
		\includegraphics[width=0.7\textwidth]{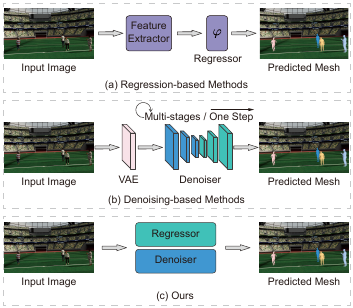}
		\caption{Comparison of our framework with existing methods for 3D human mesh recovery.} 
		\label{fig4-1}
	\end{figure}

	The regression-based paradigm has emerged as the mainstream approach for 3D human mesh recovery. 
	As illustrated in Figure \ref{fig4-1}(a), it leverages the powerful representational capacity of deep neural networks to directly regress 3D model parameters from input images in an end-to-end manner. 
	By circumventing time-consuming iterative optimization processes, this paradigm inherently excels in inference speed. 
	Furthermore, training on large-scale datasets endows it with enhanced robustness against challenges such as complex articulations.
	Despite the remarkable progress achieved by regression-based methods, occlusion remains a critical bottleneck limiting their performance and stability in unconstrained real-world scenarios. 
	In practice, the human body is frequently obscured by partial self-occlusion, human-human interactions, or scene objects. 
	Under such circumstances, the available visual cues are incomplete, causing the network to struggle to infer the accurate poses and shapes of the unseen regions. 
	Consequently, this often leads to physically implausible or geometrically erroneous estimations.

	The recent advent of denoising diffusion probabilistic models (DDPMs) offers a novel perspective to tackle the aforementioned challenges. 
	As depicted in Figure \ref{fig4-1}(b), by learning a reverse denoising process, diffusion models can generate high-quality, diverse, and structurally coherent samples. 
	Recently, the integration of conditional adapters, such as ControlNet \cite{zhang2023adding}, has empowered these models to synthesize highly controllable content via iterative denoising, guided by spatial structural cues like pose skeletons. 
	This endows them with the potential to plausibly hallucinate occluded regions even in the absence of direct visual cues. 
	Nevertheless, ControlNet-conditioned diffusion models are inherently generative. 
	When confronted with extreme or rare poses, their powerful generative priors may override the weak visual conditioning, yielding degraded outputs rather than faithfully reconstructing the authentic poses present in the input images.

\begin{figure}[htbp]
		\centering
		\includegraphics[width=0.7\textwidth]{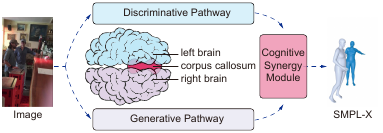}
		\caption{The framework of the proposed cognitive synergy mechanism.} 
		\label{fig4-2}
	\end{figure}

	To overcome the limitations of existing single-paradigm approaches, we propose a novel 3D human mesh recovery framework based on discriminative and generative synergy, as illustrated in Figure \ref{fig4-1}(c). 
	This framework intrinsically integrates the robust discriminative representation capacity of Vision Transformers (ViTs) with the powerful generative priors of conditional diffusion models. 
	As depicted in Figure \ref{fig4-2}, the proposed collaborative mechanism features a ViT-centric discriminative pathway. 
	Conceptually akin to the "left brain" that governs logic and analytical processing, this pathway leverages self-attention mechanisms to model global spatial dependencies across the image. 
	Consequently, it performs a process analogous to logical deduction, focusing exclusively on extracting deterministic and reliable visual cues from the visible regions of the input.
	Conversely, the generative pathway, anchored by the conditional diffusion model, functions as the "right brain" dedicated to imagination and synthesis. 
	Guided by the coarse pose skeletons injected via ControlNet, it leverages the robust generative priors of the diffusion model to synthesize physically and anatomically plausible body structures. 
	This empowers the network to plausibly hallucinate the occluded regions with high structural coherence.
	Furthermore, to maintain inference efficiency and avoid the computational bottleneck of traditional iterative sampling, this generative pathway employs a single-step diffusion denoising process. 
	This crucial design allows our framework to harness powerful generative priors without compromising the fast inference speed typical of regression-based paradigms.
	
	The remarkable cognitive capability of the human brain stems not from a mere linear combination of its left and right hemispheres, but from the intricate information exchange facilitated by the corpus callosum. 
	Analogously, in our proposed framework, this critical bridging role is undertaken by two innovative components: a Diverse-Consistent Feature Learning (DCFL) module and a Cross-Attention Multi-Level Fusion (CAMF) module. 
	Specifically, the DCFL module is designed to align the discriminative representations with the generative priors, while the cross-attention fusion mechanism enables multi-level, bidirectional feature interaction between the two pathways.

	The main contributions of this work are summarized as follows:

	(1) We propose a novel brain-inspired synergistic framework for 3D human mesh recovery, which intrinsically integrates the robust discriminative representations of ViTs with the powerful generative priors of conditional diffusion models.
	
	(2) We design a Diverse-Consistent Feature Learning (DCFL) module to effectively align the deterministic discriminative representations with the generative priors.
	
	(3) We develop a Cross-attention Multi-Level Fusion (CAMF) mechanism to facilitate bidirectional feature interaction between the two pathways across multiple semantic levels.

	(4) Extensive experiments on standard benchmark datasets demonstrate that our approach achieves superior performance across key evaluation metrics, validating its effectiveness and robustness in complex real-world scenarios.

\section{Related Work} \label{sec2}

	Since the proposed framework is founded on the synergistic integration of ViTs and diffusion models, this section provides a concise overview of regression-based paradigms and conditional diffusion models to better elucidate our methodology.

\subsection{Explicit regression methods}

	Reconstructing 3D humans from monocular images is a long-standing challenge in the computer vision community. 
	In particular, following the advent of parametric human body models \cite{anguelov2005scape, loper2023smpl, pavlakos2019expressive}, 3D human mesh recovery has garnered extensive research interest. 
	To address the inherent ambiguity of lifting 2D images to well-aligned and physically plausible 3D meshes, existing approaches have predominantly evolved into two primary paradigms. 
	The optimization-based paradigm \cite{gartner2022differentiable} typically relies on iterative fitting processes driven by various data terms and regularization constraints. 
	Conversely, the regression-based paradigm \cite{baradel2024multi, wang2025prompthmr} leverages the robust representational capacity of deep neural networks to directly regress 3D model parameters in an end-to-end manner.
	
	Currently, the regression-based paradigm dominates the field of 3D human mesh recovery. 
	By circumventing time-consuming iterative optimization processes, it facilitates real-time inference. 
	Furthermore, the rich data-driven priors acquired through deep neural networks endow this paradigm with enhanced robustness against complex articulations. 
	Nevertheless, occlusion remains a ubiquitous challenge in unconstrained real-world scenarios. 
	In practice, subjects are frequently obscured by partial self-occlusion, dense human-human interactions, or scene objects. 
	Although occlusion handling has been extensively investigated in recent years \cite{zhang2019danet, zhang2020object, yang2022lasor, zhang20233d}, the robustness and stability of existing methods still leave significant room for improvement.

\subsection{Diffusion-based methods}

	The fundamental principle of DDPMs lies in training a denoising autoencoder to estimate the reverse of a Markovian diffusion process \cite{sohl2015deep}, endowing these models with powerful generative priors. 
	Specifically, this progressive denoising procedure iteratively applies the autoencoder during sampling to synthesize high-quality, diverse, and structurally coherent data samples. 
	By formulating the recovery task as learning the reverse diffusion process of parametric human body model parameters, DDPMs can be effectively extended to tackle 3D human mesh recovery \cite{foo2023distribution, cho2023generative}. 
	
	Recently, conditional adapters tailored for diffusion models have significantly enhanced the structural and content-wise controllability of the generation process by incorporating visual conditioning. 
	Among these, while IP-Adapter \cite{ye2023ip} focuses on extracting semantic features via image prompts, ControlNet \cite{zhang2023adding} emphasizes fine-grained control through spatial structural cues, such as edge maps and pose skeletons, rendering it highly suitable for 3D human digitization. 
	Consequently, leveraging such conditional adapters can substantially augment the structural coherence of the estimated human meshes, thereby improving reconstruction precision and robustness across complex real-world scenarios \cite{zhu2024dpmesh}.

\section{Method}

	In this section, we first clarify the objectives of our learning task. 
	Then, we present the main components of the proposed framework for 3D human mesh recovery, including a diverse-consistent feature learning module, a cross-attention multi-level fusion mechanism and an instance-aware perception head.

\subsection{Preliminary}

	This work aims to tackle robust 3D human mesh recovery from a single RGB image. 
	Specifically, the 3D human body is represented using the parametric SMPL-X model, which is driven by three distinct low-dimensional vectors: pose parameters $\theta \in \mathbb{R}^{53 \times 3}$, shape parameters $\beta \in \mathbb{R}^{10}$, and facial expression parameters $\alpha \in \mathbb{R}^{10}$. 
	Consequently, the full 3D human mesh is generated via a differentiable mapping function $\mathcal{H}(\theta, \beta, \alpha) \in \mathbb{R}^{10475 \times 3}$.
	
	We formulate this task as learning a deep neural network $f_{\varPhi}$, which takes an input image $I \in \mathbb{R}^{H \times W \times 3}$ and directly regresses an independent set of SMPL-X parameters for each human instance $k$ in the scene:
	\begin{equation}
		f_{\varPhi}(I) \rightarrow \left\{ \mathcal{H}(\theta_k, \beta_k, \alpha_k) \in \mathbb{R}^{10475 \times 3} \right\}_{k=1}^{K},
		\label{eq4-1}
	\end{equation}
	where $K$ denotes the total number of detected human instances in the image.
	
	Furthermore, a pre-trained estimator is utilized to extract 2D joint coordinates $\boldsymbol{P}_{2D} \in \mathbb{R}^{N \times 2}$, which serve as explicit spatial structural cues for the reverse denoising process. 
	Inherently, diffusion models achieve a high degree of controllability via the cross-attention mechanisms embedded within their denoising U-Net architecture \cite{ronneberger2015u}. 
	By leveraging a trainable copy of the U-Net encoder, ControlNet \cite{zhang2023adding} functions as a robust conditional adapter. 
	This elegant design successfully injects the aforementioned 2D pose skeletons into the Latent Diffusion Model (LDM) \cite{rombach2022high}, significantly enhancing its fine-grained spatial controllability for structural synthesis.
	During the training phase of the conditional diffusion model, the input RGB image is first projected into a low-dimensional latent representation $\boldsymbol{z}_0$ by the encoder $E$ of a pre-trained VQ-GAN. 
	Subsequently, the noisy latent $\boldsymbol{z}_t$ at an arbitrary time step $t$ is obtained through a forward Markovian diffusion process, formulated as:
	\begin{equation}
		\boldsymbol{z}_t = \sqrt{\bar{\alpha}_t} \boldsymbol{z}_0 + \sqrt{1 - \bar{\alpha}_t} \epsilon,
		\label{eq4-2}
	\end{equation}
	where $\bar{\alpha}_t = \prod_{s=1}^t \alpha_s$, $\epsilon \sim \mathcal{N}(0, \mathbf{I})$ represents the added standard Gaussian noise, and $\mathbf{I}$ denotes the identity matrix.
	
	Given the noisy latent and the corresponding conditioning signals, the training objective of the ControlNet-conditioned latent diffusion model is formulated as a standard noise prediction loss:
	\begin{equation}
		L_{\mathrm{CLDM}} = \mathbb{E}_{\boldsymbol{z}_0, t, \boldsymbol{c}_\mathrm{t}, \boldsymbol{c}_\mathrm{f}, \boldsymbol{\epsilon}} \left[ \| \boldsymbol{\epsilon} - \boldsymbol{\epsilon}_\theta(\boldsymbol{z}_t, t, \boldsymbol{c}_\mathrm{t}, \boldsymbol{c}_\mathrm{f}) \|_2^2 \right],
		\label{eq4-3}
	\end{equation}
	where $\boldsymbol{z}_t$ is computed via Eq. \eqref{eq4-2}, $\boldsymbol{c}_\mathrm{t}$ represents the textual conditioning embeddings extracted by a pre-trained CLIP text encoder \cite{radford2021learning}, and $\boldsymbol{c}_\mathrm{f}$ denotes the task-specific spatial structural cues.
	
	In our proposed framework, the spatial conditioning signal $\boldsymbol{c}_\mathrm{f}$ is instantiated as the 2D human skeletal pose $\boldsymbol{P}_{2D}$. 
	Concurrently, the textual conditioning signal $\boldsymbol{c}_\mathrm{t}$ is derived by lifting the dimensions of the 2D joint coordinates to align with the latent dimension of the CLIP text embeddings. 
	To prevent detrimental noise from corrupting the deep hidden states during the initial training stages, ControlNet employs zero-convolution layers on the trainable adapter branch. 
	This conditional branch takes $\boldsymbol{c}_\mathrm{f}$ as input and injects the extracted hierarchical features into the corresponding output blocks of the pre-trained diffusion U-Net $\boldsymbol{\epsilon}_\theta$. 
	Crucially, to preserve the robust generative priors and alleviate computational overhead, the original parameters of $\boldsymbol{\epsilon}_\theta$ are strictly frozen. 
	Consequently, guided by the fine-grained 2D pose skeletons, the ControlNet-conditioned pathway successfully accomplishes highly controllable human feature synthesis.

\subsection{Overall Architecture}

\begin{figure}[htbp]
		\centering
		\includegraphics[width=1.0\textwidth]{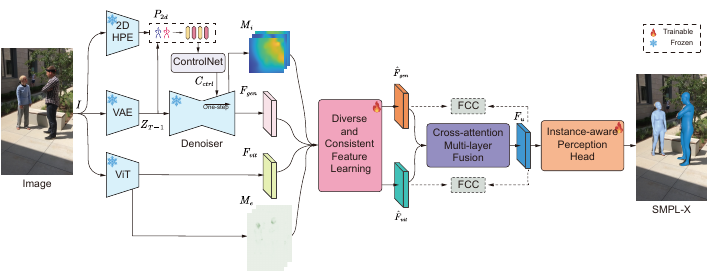}
		\caption{An overview of our proposed framework.} 
		\label{fig4-3}
	\end{figure}

	The framework proposed in this paper is illustrated in Figure \ref{fig4-3}. 
	Given an input image $I\in \mathbb{R}^{H\times W\times 3}$, it is first fed into three pre-trained networks, comprising a 2D pose estimator \cite{sun2019deep}, a Variational Autoencoder (VAE) \cite{kingma2013auto}, and a Vision Transformer-based feature extractor, DINOv2 \cite{oquab2023dinov2}. 
	Specifically, the DINOv2 feature extractor captures explicit global information from the image as the discriminative feature $F_d$, and extracts explicit intermediate attention maps $M_e$. 
	Simultaneously, the VAE encodes the input image into a compact latent space $\boldsymbol{z}_0$. 
	From this latent code, the denoiser generates the generative prior feature $F_p$ via a single-step denoising process, and identically extracts intermediate attention maps $M_i$ from the denoiser's decoder.

	The proposed dual-pathway synergistic framework is illustrated in Figure \ref{fig4-3}. 
	Given an input RGB image $I \in \mathbb{R}^{H \times W \times 3}$, it is first processed by three pre-trained modules: a 2D pose estimator \cite{sun2019deep}, a VAE \cite{kingma2013auto}, and a Vision Transformer-based discriminative encoder, DINOv2 \cite{oquab2023dinov2}. 
	Specifically, the DINOv2 encoder captures deterministic global representations from the image to serve as the discriminative feature $F_d$, concurrently extracting explicit intermediate attention maps $M_e$. 
	Simultaneously, the VAE projects the input image into a compact latent space, yielding the latent code $\boldsymbol{z}_0$. 
	Conditioned on this latent representation, the pre-trained diffusion U-Net synthesizes the robust generative prior feature $F_p$ via a single-step denoising process, while correspondingly extracting the implicit intermediate attention maps $M_i$ from its decoder layers.

	Our framework adopts a dual-pathway conditional injection strategy to introduce explicit spatial structural cues, guiding the pre-trained diffusion U-Net to attend to anatomical regions of interest. 
	Specifically, a pre-trained 2D pose estimator is leveraged to extract the 2D joint coordinates $\boldsymbol{P}_{2D}$ alongside their corresponding confidence scores. 
	For each detected human instance, 2D Gaussian kernels are applied to render a spatial heatmap $I_{h2d}$ as the dense conditional input. 
	This heatmap is channel-wise concatenated with the latent code $\boldsymbol{z}_0$ to formulate the spatial conditioning signal $\boldsymbol{c}_\mathrm{j}$ (corresponding to $\boldsymbol{c}_\mathrm{f}$ as previously denoted), which is subsequently injected into the intermediate feature maps of the denoising U-Net $\boldsymbol{\epsilon}_\theta$ via the conditional adapter. 

	Inspired by prior work \cite{zhu2024dpmesh}, the prompt conditioning signal $\boldsymbol{c}_\mathrm{t}$ circumvents the use of standard textual embeddings from a frozen CLIP model. 
	Instead, it directly utilizes the visible sparse 2D joint coordinates $\boldsymbol{P}_{2D}$. 
	A two-layer Multi-Layer Perceptron (MLP) is utilized to project these coordinates into a higher-dimensional space, strictly aligning them with the default latent dimension (i.e., 768) of the diffusion model's cross-attention mechanisms. 
	Consequently, this auxiliary spatial condition $\boldsymbol{c}_\mathrm{t}$ serves as a structural prompt, dispatched to all cross-attention layers within the U-Net. 
	Ultimately, a comprehensive conditioning set $\boldsymbol{C}_\mathrm{ctrl}$, comprising both the dense heatmap $\boldsymbol{c}_\mathrm{j}$ and the sparse prompt $\boldsymbol{c}_\mathrm{t}$, is constructed and injected into $\boldsymbol{\epsilon}_\theta$ through these distinct hierarchical pathways.

	Subsequently, the deterministic discriminative feature $F_d$, the robust generative prior feature $F_p$, and their respective intermediate attention maps $M_e$ and $M_i$ undergo synergistic alignment through the DCFL module. 
	This crucial step yields the harmonized discriminative representation $\hat{F}_d$ and generative prior $\hat{F}_p$. 
	These two aligned features are then fed into the CAMF module. 
	By employing a symmetric bidirectional cross-attention mechanism, this module effectively circumvents the inherent issues of information asymmetry and feature-level mismatch that typically arise from unidirectional information flow. 
	Finally, an instance-aware perception head leverages the human bounding masks derived from the 2D pose estimator as spatial positional priors $M_{loc}$. 
	Guided by these priors, the head decouples the fused global representation $F_u$, thereby independently regressing an accurate set of SMPL-X parameters for each distinct human instance in the complex scene.

	To train the proposed synergistic framework end-to-end, we formulate a composite objective function $\mathcal{L}_{total}$:
	\begin{equation}
		\mathcal{L}_{total} = \lambda \cdot \mathcal{L}_{align} + (1 - \lambda) \cdot \mathcal{L}_{reg},
		\label{eq4-4}
	\end{equation}
	where $\mathcal{L}_{align}$ denotes the feature consistency alignment loss enforced by the DCFL module, $\mathcal{L}_{reg}$ represents the standard regression loss supervised by the ground-truth SMPL-X parameters, and $\lambda \in [0, 1]$ is a scalar hyperparameter designed to balance the contributions of these two distinct optimization objectives.

\subsection{Diverse-Consistent Feature Learning} \label{sec3.3}	

	Conditional diffusion models leverage robust generative priors to synthesize fine-grained anatomical structures, whereas ViTs excel at extracting deterministic discriminative representations enriched with global semantics and contextual cues. 
	Given their inherently distinct representational paradigms, establishing an effective mechanism for synergistic feature alignment is of paramount importance. 
	To this end, we design the Diverse-Consistent Feature Learning (DCFL) module. This crucial component aims to harness the structural diversity of the generative pathway to augment the robustness and generalization of the discriminative features, while concurrently employing the reliable visual evidence from the ViT to rigorously guide and refine the generative priors.
	
\begin{figure}[htbp]
		\centering
		\includegraphics[width=0.45\textwidth]{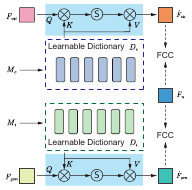}
		\caption{The structure of diverse-consistent feature learning module.} 
		\label{fig4-4}
	\end{figure}

	As illustrated in Figure \ref{fig4-4}, the proposed DCFL module primarily comprises two core components: a diverse feature representation block and learnable discriminative/generative-prior feature dictionaries. 
	The fundamental principle underlying this design is to exploit a specialized feature dictionary to compensate for the informational deficits of its counterpart representation. 
	The primary objective is to maximally preserve the intrinsic characteristics of each modality, while ensuring that either pathway can comprehensively assimilate complementary cues from the other via the injection of supplementary priors encoded within the respective dictionaries.
	
	Specifically, let $F_{vit}$ and $F_{gen}$ denote the original discriminative representation and the generative prior feature, respectively. 
	The DCFL module explicitly constructs a learnable discriminative feature dictionary $D_e$ guided by the intermediate explicit attention map $M_e$, alongside a learnable generative prior feature dictionary $D_i$ conditioned on the implicit attention map $M_i$. 
	Within the diverse feature representation block, cross-modal semantic interaction is effectively executed via standard attention mechanisms, parameterized by linear projection weight matrices ($\mathbf{W}^Q_{vit}, \mathbf{W}^K_{vit}, \mathbf{W}^V_{vit}$) and ($\mathbf{W}^Q_{gen}, \mathbf{W}^K_{gen}, \mathbf{W}^V_{gen}$). 
	Augmented by the auxiliary compensation from their counterpart dictionaries $D_i$ and $D_e$, the enhanced feature representations are denoted as $\hat{F}_{vit}$ and $\hat{F}_{gen}$, respectively. 
	Consequently, following this modality-specific dictionary compensation, $\hat{F}_{vit}$ and $\hat{F}_{gen}$ are successfully enriched with complementary missing cues, thereby substantially augmenting the comprehensive representational capacity of both pathways.
	
	Furthermore, to establish a global structural constraint, we introduce a fused representation $F_u$, which comprehensively encapsulates both the modality-specific and shared semantic cues from the discriminative and generative prior pathways. 
	Utilizing $F_u$ as a global anchor to guide the learning processes of $F_{vit}$ and $F_{gen}$ provides explicit supervisory signals, effectively mitigating the inherent limitations of single-modality representations. 
	This mutual alignment mechanism is pivotal for bridging the semantic domain gap between the latent diffusion U-Net and the Vision Transformer. 
	
	To realize this objective, the dictionary-enhanced features $\hat{F}_{vit}$ and $\hat{F}_{gen}$ are strictly aligned with the global fused feature $F_u$. 
	Specifically, we employ Feature-level Cross-Correlation (FCC) \cite{sarvaiya2009image} to explicitly quantify and maximize their spatial-channel correspondence, thereby facilitating the efficient joint optimization of the synergistic framework. Consequently, the modality-specific dictionaries $D_e$ and $D_i$ are optimized via the following alignment loss:
	\begin{equation}
		\mathcal{L}_{align} = -\operatorname{FCC}(\hat{F}_{gen}, F_u) - \operatorname{FCC}(\hat{F}_{vit}, F_u),
		\label{eq4-5}
	\end{equation}
	where the $\operatorname{FCC}$ metric between two arbitrary feature tensors $\boldsymbol{X}$ and $\boldsymbol{Y}$ is computed as:
	\begin{equation}
		\operatorname{FCC}(\boldsymbol{X}, \boldsymbol{Y}) = \frac{\sum_{i,j,k} \left(\boldsymbol{X}(i,j,k) - \bar{\boldsymbol{X}}(k)\right) \left(\boldsymbol{Y}(i,j,k) - \bar{\boldsymbol{Y}}(k)\right)}{\sqrt{\sum_{i,j,k} \left(\boldsymbol{X}(i,j,k) - \bar{\boldsymbol{X}}(k)\right)^2} \sqrt{\sum_{i,j,k} \left(\boldsymbol{Y}(i,j,k) - \bar{\boldsymbol{Y}}(k)\right)^2}}.
		\label{eq4-6}
	\end{equation}
	Here, $\boldsymbol{X}(i,j,k)$ and $\boldsymbol{Y}(i,j,k)$ denote the tensor activations at spatial location $(i,j)$ and channel $k$, while $\bar{\boldsymbol{X}}(k)$ and $\bar{\boldsymbol{Y}}(k)$ represent the spatial mean values of the $k$-th channel for tensors $\boldsymbol{X}$ and $\boldsymbol{Y}$, respectively.

\subsection{Cross-Attention Multi-Level Fusion} \label{sec3.4}

\begin{figure}[htbp]
		\centering
		\includegraphics[width=0.65\textwidth]{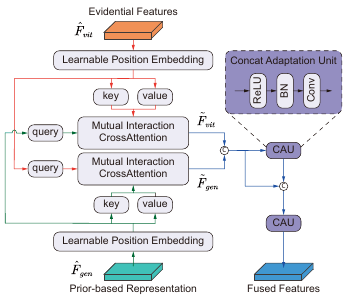}
		\caption{The details of cross-attention multi-level fusion.} 
		\label{fig4-5}
	\end{figure}

	Effectively integrating the robust generative priors with the deterministic discriminative representations is pivotal to the success of our synergistic framework. 
	Relying on naive fusion heuristics, such as direct channel concatenation or element-wise addition, not only fails to fully harness their complementary strengths but can also precipitate performance degradation due to semantic representation conflicts. 
	To mitigate this, we design the Cross-Attention Multi-Level Fusion (CAMF) module, which facilitates symmetric, bidirectional semantic interaction between the two modalities via cross-attention mechanisms. 
	The detailed architecture of this module is illustrated in Figure \ref{fig4-5}.
	
	The CAMF module ingests two parallel inputs: the aligned discriminative representation $\hat{F}_d$ and the generative prior $\hat{F}_p$. 
	To inject spatial awareness, each feature sequence is first appended with learnable positional embeddings, which automatically optimize the most suitable positional representations for the regression task during the end-to-end training process. 
	Subsequently, the bidirectional semantic interaction is formulated as:
	\begin{equation}
		\begin{cases}
			\operatorname{MICA}(\mathbf{q}_p, \mathbf{k}_d, \mathbf{v}_d) = \tilde{F}_d \\
			\operatorname{MICA}(\mathbf{q}_d, \mathbf{k}_p, \mathbf{v}_p) = \tilde{F}_p,
		\end{cases}
		\label{eq4-7}
	\end{equation}
	where $\operatorname{MICA}(\cdot)$ denotes the Multi-level Interactive Cross-Attention mechanism designed for symmetric, bidirectional semantic exchange. 
	Here, $\mathbf{q}_p$, $\mathbf{k}_p$, and $\mathbf{v}_p$ represent the projected query, key, and value tensors of the generative prior $\hat{F}_p$, whereas $\mathbf{q}_d$, $\mathbf{k}_d$, and $\mathbf{v}_d$ represent those corresponding to the discriminative representation $\hat{F}_d$.
	
	The refined representations output by the multi-level cross-attention mechanisms, $\tilde{F}_d$ and $\tilde{F}_p$, are not naively aggregated. 
	Instead, they are sequentially processed through a custom-designed Concat Adaptation Unit (CAU). 
	Structurally, each CAU comprises a robust cascade of a convolutional layer, a Batch Normalization (BN) layer, and a ReLU activation function. 
	Upon progressive integration via multiple sequential CAUs, the module ultimately synthesizes a deeply fused global representation $F_u$. 
	Crucially, this unified tensor comprehensively encapsulates the synergistic semantic cues from both the deterministic discriminative pathway and the robust generative priors, having undergone rigorous bidirectional alignment and multi-level refinement.

\subsection{Instance-Aware Perception Head} \label{sec3.5}

\begin{figure}[htbp]
		\centering
		\includegraphics[width=0.60\textwidth]{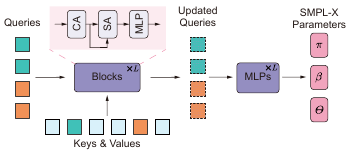}
		\caption{The details of instance-aware perception head.} 
		\label{fig4-6}
	\end{figure}

	As illustrated in Figure \ref{fig4-6}, the Instance-Aware Perception Head is fundamentally predicated on a ViT decoder architecture, comprising a sequential stack of $L$ identical multi-head attention blocks. 
	Structurally, the input Queries are initialized as a set of learnable tokens inherently encapsulating instance-specific semantics, whereas the corresponding Keys and Values are densely projected from the aforementioned fused global representation $F_u$. 
	Specifically, each ViT decoder block encompasses two primary attention mechanisms. 
	First, a self-attention module facilitates comprehensive interaction among the instance Queries, effectively modeling complex inter-instance relationships and spatial dependencies. 
	Subsequently, a cross-attention module employs these contextualized Queries to selectively scrutinize and aggregate relevant features from the dense Keys and Values of the global representation map.
		
	Driven by the cross-attention mechanism, each Query dynamically aggregates the most discriminative contextual cues pertaining to its target instance from the fused global representation $F_u$. 
	Following the computation within a single decoder block, the input Query effectively assimilates these instance-specific features, yielding an updated query embedding. 
	This sequential refinement process is iteratively repeated across all $L$ transformer layers, progressively enriching and refining the latent instance semantics encoded within the Query. 
	Ultimately, each fully contextualized query is fed into a dedicated regression head—instantiated as a MLP network—to directly regress the precise SMPL-X parameters for every distinct human instance in the scene.	

\section{Experiments}
	
	In this section, we first introduce the benchmark dataset and the evaluation metrics used to assess the proposed framework, along with the experimental setup. We then evaluate the framework and compare it with current state-of-the-art methods, presenting both quantitative and qualitative results. Finally, we conduct ablation studies to assess the impact of each proposed design.
	
\subsection{Experimental Setup}

\subsubsection{Datasets and Evaluation Metrics}

	To comprehensively evaluate the effectiveness and robustness of our proposed synergistic framework, we conduct extensive experiments across a wide array of standard benchmarks and challenging occlusion-specific datasets. 
	These include 3DPW \cite{von2018recovering}, 3DOH \cite{zhang2020object}, 3DPW-PC \cite{sun2021monocular, von2018recovering}, 3DPW-OC \cite{von2018recovering, zhang2020object}, 3DPW-Crowd \cite{choi2022learning, von2018recovering}, CMU-Panoptic \cite{joo2015panoptic}, EHF \cite{pavlakos2019expressive}, and AGORA \cite{patel2021agora}. 
	
	We adopt several standard evaluation metrics to quantitatively assess our method: Mean Per Joint Position Error (MPJPE) for measuring the absolute 3D joint accuracy, Procrustes-Aligned MPJPE (PA-MPJPE) for scale- and rotation-invariant joint evaluation, and Mean Per Vertex Error (MPVE) for assessing the fine-grained 3D mesh reconstruction quality. 
	Following the evaluation protocols established by prior works, we solely report MPJPE for the CMU-Panoptic dataset. 
	Furthermore, consistent with prevalent practices on the highly expressive EHF and AGORA benchmarks, we utilize Per-Vertex Error (PVE) to quantify the overall 3D mesh precision, while additionally reporting region-specific PVE metrics for the hands and face to highlight our model's fine-grained expressive capture capabilities.

\subsubsection{Implementation Details}

	We train our model on a hybrid dataset with 3D annotations, which includes Human3.6M \cite{ionescu2013human3}, MSCOCO \cite{lin2014microsoft}, AGORA \cite{patel2021agora}, and CrowdPose \cite{li2019crowdpose}. In this section, we utilize input images with a resolution of 448 $\times$ 448, resizing the longest edge to 448 and zero-padding the shorter edge to preserve the aspect ratio. The only data augmentation employed is random horizontal flipping. Additionally, we adopt the neutral SMPL-X model parameterized by 10 shape components. The proposed model is trained for 400,000 iterations using Automatic Mixed Precision (AMP) and the AdamW optimizer \cite{loshchilov2017decoupled}, with a batch size of 8 and a weight decay of $1e^{-6}$. The initial learning rate is set to $1e^{-4}$ and subsequently decayed to $1e^{-5}$ during the final 5 epochs.

\subsection{Results and Analysis}

\subsubsection{Quantitative and Qualitative Results}

	\textbf{3DPW-OC} is a subset of the 3DPW dataset featuring person-object occlusion, containing 20,243 individuals. For a fair comparison, all methods reported in this paper are evaluated without fine-tuning on its training set. As shown in Table \ref{tab4-1}, our model significantly outperforms all other compared methods. Compared to VMarker-Pro, it reduces the MPJPE by 3.5 mm (4.9\%), decreases the PA-MPJPE by 1.9 mm, and achieves a 3.4 mm reduction in the more fine-grained MPVE metric.
	
	\textbf{3DPW-PC} is a subset of the 3DPW dataset featuring person-person occlusion, containing 2,218 individuals. As shown in Table \ref{tab4-1}, our method reduces the MPJPE and PA-MPJPE by 2.2 mm and 5.4 mm, respectively, compared to DPMesh.

\begin{table}[htbp]
	\centering
	\caption{Performance comparison of the proposed method on the 3DPW-OC and 3DPW-PC datasets (in mm).}
	\resizebox{0.9\textwidth}{!}{
	\begin{tabular}{lcccccc}
		\toprule[1pt]
		\multirow{2}{*}{} & \multicolumn{3}{c|}{3DPW-OC} & \multicolumn{3}{c}{3DPW-PC} \\
		& MPJPE $\downarrow$  & PA-MPJPE $\downarrow$  & \multicolumn{1}{c|}{MPVE $\downarrow$}   & MPJPE $\downarrow$  & PA-MPJPE $\downarrow$  & MPVE $\downarrow$   \\ \midrule[0.75pt]
		SPIN \cite{kolotouros2019learning}                   & 95.5   & 60.7      & 121.4  & 122.1  & 77.5      & 159.8  \\
		PyMAF \cite{zhang2021pymaf}                  & 89.6   & 59.1      & 113.7  & 117.5  & 74.5      & 154.6  \\
		ROMP \cite{sun2021monocular}                   & 91.0     & 66.5      & -      & 98.7   & 75.8      & -      \\
		OCHMR \cite{khirodkar2022occluded}                  & 112.2  & 75.2      & 145.9  & 105.8  & 72.6      & 142.0    \\
		PARE \cite{kocabas2021pare}                   & 90.5   & 56.6      & 107.9  & 95.8   & 64.5      & 122.4  \\
		3DCrowdNet \cite{choi2022learning}             & 83.5   & 57.1      & 101.5  & 90.9   & 64.4      & 114.8  \\
		VMarker-Pro \cite{ma2025vmarker}            & 70.7   & 45.1      & 84.8   & -      & -         & -      \\
		DPMesh \cite{zhu2024dpmesh}                 & 70.9   & 48.0        & 88.0     & 82.2   & 56.6      & 105.4  \\
		\cellcolor[rgb]{.851,.851,.851}Ours                    & \cellcolor[rgb]{.851,.851,.851}$\mathbf{67.2}$   & \cellcolor[rgb]{.851,.851,.851}$\mathbf{43.2}$      & \cellcolor[rgb]{.851,.851,.851}$\mathbf{81.4}$   & \cellcolor[rgb]{.851,.851,.851}$\mathbf{80.0}$     & \cellcolor[rgb]{.851,.851,.851}$\mathbf{51.2}$      & \cellcolor[rgb]{.851,.851,.851}$\mathbf{101.7}$  \\ \bottomrule[1pt]
	\end{tabular}
	}		
	\label{tab4-1}
	\end{table}	

	\textbf{3DOH} is a dataset specifically designed for person-object occlusion, with a test set comprising 1,290 images. 
	As shown in Table \ref{tab4-2}, although the PARE method achieves prominent performance in terms of PA-MPJPE, our proposed method remains highly competitive, yielding an MPJPE of 80.9 mm.
	
	\textbf{3DPW-Crowd} is a densely crowded subset of the 3DPW dataset, encompassing 1,923 individuals. As shown in Table \ref{tab4-2}, our model achieves the state-of-the-art results. 
	Compared to DPMesh, it reduces the MPJPE and PA-MPJPE by 4.5 mm (a relative improvement of 5.6\%) and 3.0 mm (a relative improvement of 5.9\%), respectively, while also decreasing the MPVE by 3.2 mm.

	\begin{table}[htbp]
		\centering
		\caption{Performance comparison of the proposed method on the 3DOH and 3DPW-Crowd datasets (in mm, $\mathbf{bold}$ denotes the best result, while \underline{underline} denotes the second best).}
		\resizebox{0.95\textwidth}{!}{
		\begin{tabular}{lcccccc}
			\toprule[1pt]
			\multirow{2}{*}{} & \multicolumn{3}{c|}{3DOH} & \multicolumn{3}{c}{3DPW-Crowd} \\
			& MPJPE $\downarrow$ & PA-MPJPE $\downarrow$ & \multicolumn{1}{c|}{MPVE $\downarrow$}  & MPJPE $\downarrow$   & PA-MPJPE $\downarrow$   & MPVE $\downarrow$    \\ \midrule[0.75pt]
			SPIN \cite{kolotouros2019learning}             & 110.5 & 71.6     & 124.2 & 121.2   & 69.9       & 144.1   \\
			PyMAF \cite{zhang2021pymaf}            & 96.2  & 67.7     & 107.3 & 115.7   & 66.4       & 147.5   \\
			ROMP \cite{sun2021monocular}              & -     & -        & -     & 104.8   & 63.9       & 127.8   \\
			PARE \cite{kocabas2021pare}             & $\mathbf{63.3}$  & $\mathbf{44.3}$     & -     & 94.9    & 57.5       & 117.6   \\
			3DCrowdNet \cite{choi2022learning}       & 102.8 & 61.6     & 111.8 & 85.8    & 55.8       & 108.5   \\
			DPMesh \cite{zhu2024dpmesh}           & 97.1  & 59.0       & \underline{106.4} & \underline{79.9}    & \underline{51.1}       & \underline{101.5}   \\
			\cellcolor[rgb]{.851,.851,.851}Ours              & \cellcolor[rgb]{.851,.851,.851}\underline{80.9}  & \cellcolor[rgb]{.851,.851,.851}\underline{57.2}    & \cellcolor[rgb]{.851,.851,.851}$\mathbf{101.5}$ & \cellcolor[rgb]{.851,.851,.851}$\mathbf{75.4}$    & \cellcolor[rgb]{.851,.851,.851}$\mathbf{48.1}$       & \cellcolor[rgb]{.851,.851,.851}$\mathbf{98.3}$   \\ \bottomrule[1pt]
		\end{tabular}
		}
		\label{tab4-2}
	\end{table}

\begin{figure}[htbp]
		\centering
		\includegraphics[width=1\textwidth]{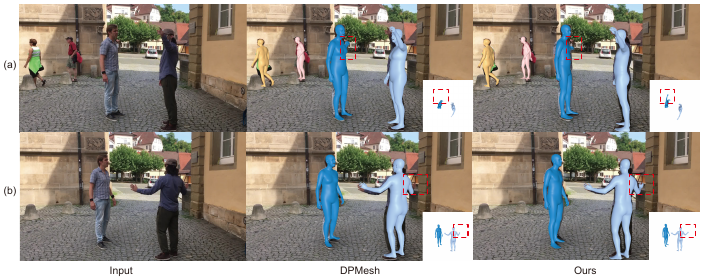}
		\caption{Qualitative comparison of experimental results on the 3DPW dataset.} 
		\label{fig4-8}
	\end{figure}

	\textbf{3DPW} serves as a widely adopted benchmark for 3D human reconstruction. 
	Its test set comprises 60 videos containing 3D annotations for 35,515 individuals. 
	We conduct extensive comparisons against a series of representative methods on the mainstream 3DPW test set. 
	As shown in Table \ref{tab4-3}, our method achieves state-of-the-art performance on this test set. 
	Compared to the recent VMarker-Pro, our method achieves a 2.5 mm reduction in MPJPE (a 4.4\% relative error reduction), decreases the PA-MPJPE by 1.2 mm (a 3.4\% relative reduction), and yields a 2.0 mm improvement in MPVE (a 3.1\% relative reduction). 
	When compared with PromptHMR, the performance gains are even more substantial: the MPJPE, PA-MPJPE, and MPVE are reduced by 5.0 mm, 2.8 mm, and 6.7 mm, corresponding to relative error reductions of 8.5\%, 7.7\%, and 9.7\%, respectively. 
	Qualitative results are presented in Figure \ref{fig4-8}, demonstrating the strong robustness of our model in complex scenes.

	\begin{table}[htbp]
		\centering
		\caption{Performance comparison of the proposed method on the 3DPW dataset (in mm).}
		\resizebox{0.68\textwidth}{!}{
		\begin{tabular}{lcccc}
			\toprule[1pt]
			\multirow{2}{*}{} & \multirow{2}{*}{} & \multicolumn{3}{c}{3DPW   test} \\ \cmidrule[0.75pt]{3-5}
			&                              & MPJPE $\downarrow$   & PA-MPJPE $\downarrow$  & MPVE $\downarrow$   \\ \midrule[0.75pt]
			HMR \cite{kanazawa2018end}                    & CVPR'18                      & 130.0      & 76.7       & -       \\
			GraphCMR \cite{kolotouros2019convolutional}               & CVPR'19                      & -        & 70.2       & -       \\
			SPIN \cite{kolotouros2019learning}                   & ICCV'19                      & 96.9     & 59.2       & 116.4   \\
			ROMP \cite{sun2021monocular}                   & ICCV'21                      & 76.7     & 47.3       & 93.4    \\
			PARE \cite{kocabas2021pare}                   & ICCV'21                      & 82.9     & 52.3       & 99.7    \\
			BEV \cite{sun2022putting}                    & CVPR'22                      & 78.5     & 46.9       & 92.3    \\
			Hand4Whole \cite{moon2022accurate}             & CVPR'22                      & 86.6     & 54.4       & -       \\
			3DCrowdNet \cite{choi2022learning}             & CVPR'22                      & 81.7     & 51.5       & 98.3    \\
			OCHMR \cite{khirodkar2022occluded}                  & CVPR'22                      & 89.7     & 58.3       & 107.1   \\
			PSVT \cite{qiu2023psvt}                   & CVPR'23                      & 75.5     & 45.7       & 84.9    \\
			OSX \cite{lin2023one}                    & CVPR'23                      & 74.7     & 45.1       & -       \\
			SMPLer-X \cite{cai2023smpler}               & NeurIPS'23                   & 78.3     & 52.1       & -       \\
			PyMAF \cite{zhang2021pymaf}                  & TPAMI'23                     & 92.8     & 58.9       & 110.1   \\
			DPMesh \cite{zhu2024dpmesh}                 & CVPR'24                      & 73.6     & 47.4       & 90.7    \\
			Multi-HMR \cite{baradel2024multi}              & ECCV'24                      & 61.4     & 41.7       & 75.9    \\
			PromptHMR \cite{wang2025prompthmr}              & CVPR'25                      & 58.7     & 36.6       & 69.4    \\
			VMarker-Pro \cite{ma2025vmarker}             & TPAMI'25                     & 56.2     & 35.0         & 64.7    \\
			\cellcolor[rgb]{.851,.851,.851}Ours                    &   \cellcolor[rgb]{.851,.851,.851}                           & \cellcolor[rgb]{.851,.851,.851}$\mathbf{53.7}$     & \cellcolor[rgb]{.851,.851,.851}$\mathbf{33.8}$       & \cellcolor[rgb]{.851,.851,.851}$\mathbf{62.7}$    \\ \bottomrule[1pt]
		\end{tabular}
		}
		\label{tab4-3}
	\end{table}

	\textbf{EHF} is the first dataset used for evaluating SMPL-X-based models. Constructed via a scanning system and subsequently fitted with SMPL-X meshes, it is a single-person full-body pose dataset consisting of 100 images. 
	As shown in Table \ref{tab4-4}, in terms of the PVE metric, our model achieves an overall error of 40.5 mm, which is a reduction of 1.5 mm (a relative improvement of 3.6\%) compared to Multi-HMR. 
	For the PA-PVE metric, the overall error is reduced from 28.2 mm to 27.3 mm. 
	Notably, the model excels particularly in the reconstruction of fine-grained regions such as the hands and face, decreasing the hand PVE by 1.7 mm (a relative improvement of 5.9\%) and the face PVE by 0.9 mm, while achieving state-of-the-art performance on PA-PVE for both regions.
	
\begin{figure}[htbp]
		\centering
		\includegraphics[width=1\textwidth]{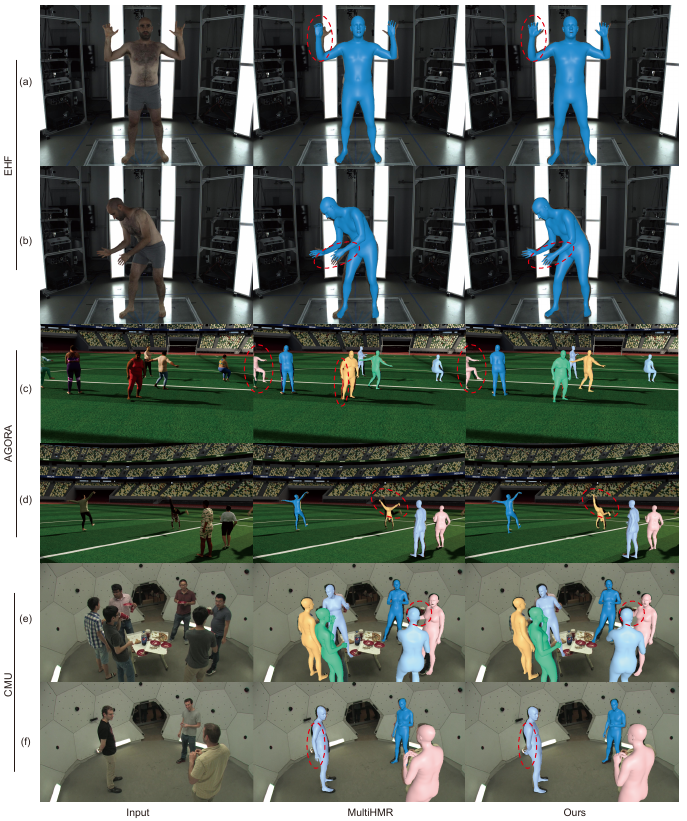}
		\caption{Qualitative comparison of experimental results on the EHF, AGORA, and CMU datasets.} 
		\label{fig4-7}
	\end{figure}

	\textbf{AGORA} is a highly realistic multi-person synthetic dataset, containing 14,000 training images, 2,000 validation images, and 3,000 test images. 
	It consists of 4,240 high-quality human scans, each equipped with precise SMPL and SMPL-X annotations. 
	The results on the test set are obtained through an online leaderboard that provides evaluations for both SMPL and SMPL-X. 
	Since the leaderboard does not provide distance estimation metrics on the test set, we follow the protocol of previous works and concurrently report the distance estimation results on the validation set. 
	As shown in Table \ref{tab4-4}, on the AGORA dataset, our method achieves a 3.1 mm improvement in overall PVE compared to Multi-HMR. 
	This improvement is primarily attributed to a substantial breakthrough in the reconstruction accuracy of the hand regions, reducing the hand PVE to 34.0 mm and realizing a performance gain of at least 5.3 mm (a relative improvement of 13.5\%). 
	Meanwhile, the face PVE also surpasses previous methods with a score of 26.4 mm. 
	We further present qualitative results on both the EHF and AGORA datasets in Figure \ref{fig4-7} (a)-(d), demonstrating the model's proficiency in handling severe occlusions.

\begin{table}[htbp]
	\centering
	\caption{Performance comparison of the proposed method on the EHF and AGORA datasets (in mm, $\mathbf{bold}$ denotes the best result, while \underline{underline} denotes the second best).}
	\resizebox{0.8\textwidth}{!}{
	\begin{tabular}{lccccccccc}
		\toprule[1pt]
		\multirow{3}{*}{} & \multicolumn{6}{c|}{EHF}                                                             & \multicolumn{3}{c}{AGORA} \\ \cmidrule[0.75pt]{2-10} 
		& \multicolumn{3}{c|}{PVE $\downarrow$}                  & \multicolumn{3}{c|}{PA-PVE $\downarrow$}              & \multicolumn{3}{c}{PVE $\downarrow$}   \\ \cmidrule[0.75pt]{2-10} 
		& All   & Hands & \multicolumn{1}{c|}{Face} & All  & Hands & \multicolumn{1}{c|}{Face} & All     & Hands   & Face  \\ \midrule[0.75pt]
		ExPose \cite{choutas2020monocular}           & 77.1  & 51.6  & 35.0                        & 54.5 & 12.8  & 5.8                       & 217.3   & 73.1    & 51.1  \\
		FrankMocap \cite{rong2021frankmocap}       & 107.6 & 42.8  & -                         & 57.5 & 12.6  & -                         & -       & 55.2    & -     \\
		PIXIE \cite{feng2021collaborative}            & 88.2  & 42.8  & 32.7                      & 55.0   & 11.1  & $\mathbf{4.6}$                       & 191.8   & 49.3    & 50.2  \\
		Hand4Whole \cite{moon2022accurate}       & 76.8  & 39.8  & 26.1                      & 50.3 & 10.8  & 5.8                       & 135.5   & 47.2    & 41.6  \\
		PyMAF-X \cite{zhang2023pymaf}           & 64.9  & 29.7  & 19.7                      & 50.2 & $\mathbf{10.2}$  & 5.5                       & 125.7   & 45.0      & 35.0    \\
		OSX \cite{lin2023one}               & 70.8  & 53.7  & 26.4                      & 48.7 & 15.9  & 6.0                         & 122.8   & 45.7    & 36.2  \\
		SMPLer-X \cite{cai2023smpler}          & 65.4  & 49.4  & \underline{17.4}                      & 37.8 & 15.0    & 5.1                       & 99.7    & \underline{39.3}    & 29.9  \\
		Multi-HMR \cite{baradel2024multi}        & \underline{42.0}    & \underline{28.9}  & 18.0                        & \underline{28.2} & 10.8  & 5.3                       & \underline{95.9}    & 40.7    & \underline{27.7}  \\
		\cellcolor[rgb]{.851,.851,.851}Ours              & \cellcolor[rgb]{.851,.851,.851}$\mathbf{40.5}$  & \cellcolor[rgb]{.851,.851,.851}$\mathbf{27.2}$  & \cellcolor[rgb]{.851,.851,.851}$\mathbf{17.1}$                      & \cellcolor[rgb]{.851,.851,.851}$\mathbf{27.3}$ & \cellcolor[rgb]{.851,.851,.851}\underline{10.6}  & \cellcolor[rgb]{.851,.851,.851}\underline{4.9}                       & \cellcolor[rgb]{.851,.851,.851}$\mathbf{92.8}$    & \cellcolor[rgb]{.851,.851,.851}$\mathbf{34.0}$      & \cellcolor[rgb]{.851,.851,.851}$\mathbf{26.4}$  \\ \bottomrule[1pt]
	\end{tabular}
	}
	\label{tab4-4}
\end{table}

	\textbf{CMU-Panoptic} is an indoor multi-person dataset captured using multi-view camera setups. 
	To ensure a fair comparison, we follow the protocol of previous works and evaluate our method on four selected scenes. 
	As shown in Table \ref{tab4-5}, our method outperforms competitors on all selected video sequences. 
	Although Multi-HMR reports a lower overall MPJPE on this dataset, our method achieves an MPJPE of 105.2 mm and exhibits robust and superior performance across specific individual scenes. 
	We further present qualitative results in Figure \ref{fig4-7} (e)-(f). 
	
	On the AGORA dataset, our method achieves an F1 score of 0.95, matching the state-of-the-art Multi-HMR. 
	However, in terms of core accuracy metrics, our model demonstrates superior performance. 
	Specifically, our method reduces the MPJPE from 65.3 mm to 61.1 mm, achieving a significant error reduction of 4.2 mm (a relative improvement of 6.4\%). 
	Simultaneously, on the PVE metric, which measures the conformity of the 3D mesh model, we also achieve an improvement from 61.1 mm to 58.4 mm.

	\begin{table}[htbp]
		\centering
		\caption{Performance comparison of the proposed method on the CMU-Panoptic and AGORA datasets (in mm, $\mathbf{bold}$ denotes the best result, while \underline{underline} denotes the second best).}
		\resizebox{0.95\textwidth}{!}{
		\begin{tabular}{lccccccccc}
			\toprule[1pt]
			\multirow{2}{*}{} & \multicolumn{6}{c|}{CMU-Panoptic}                                            & \multicolumn{3}{c}{AGORA} \\ \cmidrule[0.75pt]{2-10}
			& F1(\%) $\uparrow$  & Haggl. & Mafia & Ultim. & Pizza & \multicolumn{1}{c|}{MPJPE $\downarrow$} & F1(\%) $\uparrow$    & MPJPE $\downarrow$   & PVE $\downarrow$    \\ \midrule[0.75pt]
			CRMH \cite{jiang2020coherent}                   & 0.92 & 129.6  & 133.5 & 153    & 156.7 & 143.2                      & -      & -       & -      \\
			3DCrowdNet \cite{choi2022learning}             & -    & 109.6  & 135.9 & 129.8  & 135.6 & 127.6                      & -      & -       & -      \\
			ROMP \cite{sun2021monocular}                   & \underline{0.93} & 107.8  & 125.3 & 135.4  & 141.8 & 127.6                      & 0.91   & 108.1   & 103.4  \\
			BEV \cite{sun2022putting}                     & $\mathbf{0.97}$ & \underline{90.7}   & \underline{103.7} & \underline{113.1}  & 125.2 & 109.5                      & \underline{0.93}   & 105.3   & 100.7  \\
			Multi-HMR \cite{baradel2024multi}              & $\mathbf{0.97}$ & -      & -     & -      & -     & $\mathbf{77.3}$                      & $\mathbf{0.95}$   & \underline{65.3}    & \underline{61.1}   \\
			DPMesh \cite{zhu2024dpmesh}                 & -    & 97.2   & 109.8 & 114.3  & \underline{120.5} & 110.4                      & -      & -       & -      \\
			\cellcolor[rgb]{.851,.851,.851}Ours                    & \cellcolor[rgb]{.851,.851,.851}$\mathbf{0.97}$ & \cellcolor[rgb]{.851,.851,.851}$\mathbf{89.6}$   & \cellcolor[rgb]{.851,.851,.851}$\mathbf{101.3}$ & \cellcolor[rgb]{.851,.851,.851}$\mathbf{110.5}$  & \cellcolor[rgb]{.851,.851,.851}$\mathbf{119.6}$ & \cellcolor[rgb]{.851,.851,.851}\underline{105.2}                      & \cellcolor[rgb]{.851,.851,.851}$\mathbf{0.95}$   & \cellcolor[rgb]{.851,.851,.851}$\mathbf{61.1}$    & \cellcolor[rgb]{.851,.851,.851}$\mathbf{58.4}$   \\ \bottomrule[1pt]
		\end{tabular}
		}
		\label{tab4-5}
	\end{table}

\begin{figure}[htbp]
		\centering
		\includegraphics[width=0.9\textwidth]{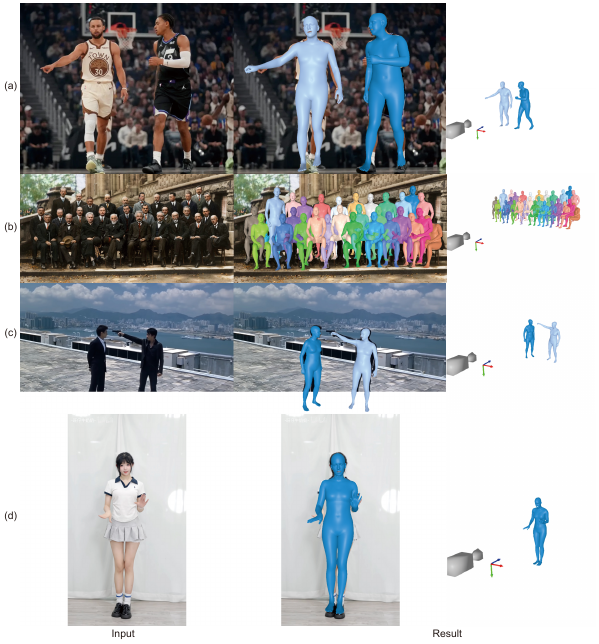}
		\caption{Visualization of Experimental Results in Real-world Scenes.} 
		\label{fig4-9}
	\end{figure}

	Estimating SMPL-X mesh parameters from videos captured in unconstrained real-world environments poses a significant challenge due to complex environmental factors and unknown camera parameters. 
	A practical approach to evaluate the generalization ability of a model is to apply the pre-trained network to such videos. 
	During the evaluation phase, we utilize the trained model to estimate SMPL-X mesh parameters in real-world scenes sourced from Bilibili. 
	A variety of challenging scenes are selected, covering basketball, dancing, movies, and group photos. 
	As illustrated in Figure \ref{fig4-9}, our method generates plausible and high-fidelity mesh results on these real-world images, demonstrating its exceptional generalization capabilities.
	
\subsubsection{Ablation Studies}

\noindent\textit{1) Impact of Different 2D Pose Estimation Pre-trained Models}

	Our framework relies on a pre-trained 2D pose estimator to obtain 2D joints $\boldsymbol{P}_{2D}$ and their confidences, which are subsequently utilized to construct conditioning injections ($c_j$ and $c_t$) to guide the diffusion model. Table \ref{tab4-6} demonstrates the impact of different 2D pose estimation backbone networks on the final 3D reconstruction accuracy. Experimental results indicate that when employing HRNet as the backbone network, the model achieves an MPJPE of 53.7 mm on 3DPW-test and a PVE of 92.8 mm on AGORA, both significantly outperforming those of ResNet-50 and ResNet-101. This is primarily attributed to HRNet's ability to consistently maintain high-resolution feature representations throughout the network, thereby providing more accurate 2D keypoint localization and spatial topological information. A more precise $\boldsymbol{P}_{2D}$ can generate higher-quality heatmaps $I_{h2d}$ and spatial prompt conditions $c_t$, enabling the denoising U-Net to focus more accurately on the target human body regions in multi-person scenes.
	
	\begin{table}[htbp]
		\centering
		\caption{Comparison of different pre-trained 2D pose estimation models (in mm).}
		\resizebox{0.6\textwidth}{!}{
			\begin{tabular}{lcc}
				\toprule[1pt]
				Pre-trained 2D Pose Estimator & \begin{tabular}[c]{@{}c@{}}3DPW-test\\ MPJPE\end{tabular} & \begin{tabular}[c]{@{}c@{}}AGORA\\ PVE\end{tabular} \\ \midrule[0.75pt]
				ResNet-50 \cite{he2016deep}       & 64.7                                                      & 117.7                                               \\
				ResNet-101 \cite{he2017mask}      & 59.3                                                      & 105.6                                               \\
				HRNet \cite{sun2019deep}           & $\mathbf{53.7}$                                                      & $\mathbf{92.8}$                                                \\ \bottomrule[1pt]
			\end{tabular}
		}
		\label{tab4-6}
	\end{table}

	\noindent\textit{2) Impact of Different Pre-trained Feature Extractors}
	
	The feature extractor is responsible for extracting explicit global perceptual information (discriminative features $F_{vit}$) from the image. 
	In Table \ref{tab4-7}, we compare two Vision Transformer-based backbone networks: DINOv2 and DeiT. 
	The data indicates that DINOv2 achieves an MPJPE of 53.7 mm, which is an error reduction of approximately 16.9\% compared to DeiT (64.6 mm). 
	Since DINOv2 is pre-trained on large-scale datasets using self-supervised learning, the features it extracts contain richer fine-grained visual cues and a stronger capability for instance discrimination. 
	This is crucial for feature alignment and decoupling in multi-person occlusion scenarios.
	
	\begin{table}[htbp]
		\centering
		\caption{Comparison of different pre-trained feature extractors (in mm).}
		\resizebox{0.45\textwidth}{!}{
			\begin{tabular}{lc}
				\toprule[1pt]
				Pre-trained Feature Extractors & \begin{tabular}[c]{@{}c@{}}3DPW-test\\ MPJPE\end{tabular} \\ \midrule[0.75pt]
				DeiT \cite{touvron2021training}          & 64.6                                                      \\ 
				DINOv2 \cite{oquab2023dinov2}        & $\mathbf{53.7}$                                                    \\ \bottomrule[1pt]
			\end{tabular}
		}
		\label{tab4-7}
	\end{table}

	\noindent\textit{3) Importance of Intermediate Attention Layers $M_e$ and $M_i$}
	
	In this chapter, we propose extracting intermediate attention layers $M_e$ and $M_i$ from the feature extractor and the denoiser respectively to assist in diverse consistency feature learning. 
	As shown in Table \ref{tab4-8}, when both are removed, the MPJPE reaches 60.8 mm. 
	When solely introducing $M_e$ or $M_i$, the error decreases to 59.6 mm and 57.9 mm, respectively. When both collaborate, the MPJPE achieves an optimal 53.7 mm. 
	This validates that the attention of the discriminative model (focusing on global context and instance boundaries) and the attention of the generative model (focusing on structural priors and local textures) are highly complementary. 
	Introducing these two intermediate attention layers effectively guides the network to filter background noise, enabling feature learning to focus more precisely on the non-rigid deformation of the human body itself.
	
	\begin{table}[htbp]
		\centering
		\caption{Ablation study of intermediate attention layers (in mm).}
		\resizebox{0.3\textwidth}{!}{
			\begin{tabular}{ccc}
				\toprule[1pt]
				$M_e$  & $M_i$  & \begin{tabular}[c]{@{}c@{}}3DPW-test\\  MPJPE\end{tabular} \\ \midrule[0.75pt]
				w   & w/o & 59.6                                                       \\
				w/o & w/o & 60.8                                                       \\
				w/o & w   & 57.9                                                       \\
				w   & w   & $\mathbf{53.7}$                                                      \\ \bottomrule[1pt]
			\end{tabular}
		}
		\label{tab4-8}
	\end{table}

	\noindent\textit{4) Impact of Different Fusion Methods for Prior Generative Features and Discriminative Features}
		
	How to effectively fuse discriminative features $\hat{F}_{vit}$ and generative prior features $\hat{F}_{gen}$ is a core focus of this chapter's research. 
	Table \ref{tab4-9} compares simple addition (Sum), channel concatenation (Concat), and the proposed Cross-Attention Multi-Level Fusion (CMF). 
	The results show that simple addition (71.1 mm) or concatenation (74.9 mm) perform poorly, even deteriorating the original feature representations. 
	Since $\hat{F}_{vit}$ and $\hat{F}_{gen}$ originate from pre-trained spaces with distinct representation mechanisms, significant semantic gaps and hierarchical mismatches exist between them. 
	The CMF module designed in this chapter, through a symmetric bidirectional cross-attention mechanism, achieves dynamic alignment and adaptive fusion of features at different semantic levels. 
	This effectively prevents information asymmetry caused by unidirectional information flow, resulting in optimal reconstruction performance on both the 3DPW and AGORA datasets (53.7 mm and 92.8 mm, respectively).
	
	\begin{table}[htbp]
		\centering
		\caption{Comparison of different feature fusion methods (in mm).}
		\resizebox{0.35\textwidth}{!}{
			\begin{tabular}{lcc}
				\toprule[1pt]
				& \begin{tabular}[c]{@{}c@{}}3DPW-test\\  MPJPE\end{tabular} & \begin{tabular}[c]{@{}c@{}}AGORA\\ PVE\end{tabular} \\ \midrule[0.75pt]
				Sum    & 71.1                                                       & 137.5                                               \\
				Concat & 74.9                                                       & 126.1                                               \\
				\textbf{CMF}    & $\mathbf{53.7}$                                                       & $\mathbf{92.8}$                                               \\ \bottomrule[1pt]
			\end{tabular}
		}
		\label{tab4-9}
	\end{table}	
	
	\subsubsection{Visualization of Cross-Attention Maps}
		
\begin{figure}[htbp]
		\centering
		\includegraphics[width=0.8\textwidth]{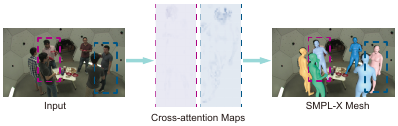}
		\caption{Visualization of Cross-Attention Maps.} 
		\label{fig4-12}
	\end{figure}

	As illustrated in Figure \ref{fig4-12}, in the given examples where the target is partially occluded, the results indicate that our method accurately captures spatial relationships and distinguishes between the two targets in the cross-attention maps, even when one target is occluded by the other. 
	This enables the model to effectively recover the occluded meshes. 
	The network model of our discriminative and generative collaborative mechanism accurately captures occlusion-aware information in the cross-attention maps. 
	By identifying targets from various occlusions, it proves that the model successfully extracts occlusion-aware knowledge, providing explicit guidance for the subsequent regressor.
		
\section{Conclusion}

	In this paper, we introduced a novel brain-inspired synergistic framework that integrates the robust discriminative representations of vision transformers with the powerful generative priors of conditional diffusion models. 
	It utilizes a collaborative cognitive mechanism to enhance 3D human mesh recovery from monocular RGB images, particularly in scenarios with severe occlusion. 
	Specifically, the ViT-centric discriminative pathway acts to extract deterministic visual cues from visible regions, while the diffusion-based generative pathway functions to synthesize anatomically plausible body structures. 
	To bridge these two pathways, the diverse-consistent feature learning module aligns discriminative representations with generative priors, and the cross-attention multi-level fusion module facilitates bidirectional feature interaction across multiple semantic levels. 
	Experiments on standard benchmark datasets affirm the efficacy and superior robustness of our proposed model compared to state-of-the-art techniques in complex real-world scenarios.
	
	\textbf{Limitations and future work.} 
	Despite the promising performance of the proposed dual-pathway synergistic framework, several limitations remain to be addressed in future research. 
	First, the architecture entails a substantial computational and memory overhead. The framework requires the concurrent execution of multiple large-scale pre-trained models, including a 2D pose estimator, a VAE, DINOv2, and a diffusion U-Net. Although we employ a single-step denoising process to accelerate the generative pathway, the overall parameter count and inference latency remain high. This complexity poses a challenge for deployment on resource-constrained edge devices or in strict real-time applications. 
	Second, the robustness of our conditional injection strategy is inherently bottlenecked by the performance of the off-the-shelf 2D pose estimator. The framework relies heavily on the detected 2D joint coordinates to formulate both the spatial heatmap and the structural prompt. In highly challenging scenarios characterized by severe inter-person occlusions, truncation, or extreme motion blur, the 2D pose estimator may yield inaccurate or missing detections. These initial errors can introduce noisy conditioning signals into the diffusion U-Net, which may subsequently propagate through the diverse-consistent feature learning and cross-attention multi-level fusion modules, ultimately degrading the accuracy of the final SMPL-X parameter regression. 
	Future work will focus on designing lightweight, distilled variants of the dual-pathway architecture to improve computational efficiency. Additionally, we aim to explore more robust, self-correcting mechanisms for spatial conditioning that can seamlessly handle noisy or incomplete 2D pose priors in heavily occluded scenes.




\small
\bibliographystyle{elsarticle-num}
\bibliography{refs.bib}





\end{document}